\documentclass{article}

\usepackage[final]{nips_2017}

\usepackage[utf8]{inputenc} 
\usepackage[T1]{fontenc}    
\usepackage{hyperref}       
\usepackage{url}            
\usepackage{booktabs}       
\usepackage{amsfonts}       
\usepackage{nicefrac}       
\usepackage{microtype}      

\usepackage{graphicx, color, xspace}
\usepackage{caption}
\usepackage{subcaption}

\title{CycleGAN, a Master of Steganography}

\usepackage{amsmath}
\def\E{\mathbb{E}}
\DeclareMathOperator*{\argmin}{arg\,min}

\author{
  Casey Chu \\
  Stanford University\\
  \texttt{caseychu@stanford.edu} \\
  \And
  Andrey Zhmoginov \\
  Google Inc.\\
  \texttt{azhmogin@google.com} \\
  \And
  Mark Sandler \\
  Google Inc.\\
  \texttt{sandler@google.com} \\
}

\newif\ifpaper

\begin{document}

\maketitle

\begin{abstract}
  CycleGAN \citep{CycleGAN2017} is one recent successful approach to learn a transformation between two image distributions. In a series of experiments, we demonstrate an intriguing property of the model: CycleGAN learns to ``hide'' information about a source image into the images it generates in a nearly imperceptible, high-frequency signal. This trick ensures that the generator can recover the original sample and thus satisfy the cyclic consistency requirement, while the generated image remains realistic. We connect this phenomenon with adversarial attacks by viewing CycleGAN's training procedure as training a generator of adversarial examples and demonstrate that the cyclic consistency loss causes CycleGAN to be especially vulnerable to adversarial attacks.
\end{abstract}

\section{Introduction}
Image-to-image translation is the task of taking an image from one class of images and rendering it in the style of another class. One famous example is artistic style transfer, pioneered by \cite{gatys2015neural}, which is the task of rendering a photograph in the style of a famous painter.

One recent technique for image-to-image translation is CycleGAN \citep{CycleGAN2017}. It is particularly powerful because it requires only \emph{unpaired} examples from two image domains $X$ and $Y$. CycleGAN works by training two transformations $F: X \to Y$ and $G: Y \to X$ in parallel, with the goal of satisfying the following two conditions:
\begin{enumerate}
    \item $Fx \sim p(y)$ for $x\sim p(x)$, and $Gy \sim p(x)$ for $y\sim p(y)$;
    \item $GFx = x$ for all $x\in X$, and $FGy = y$ for all $y\in Y$,
\end{enumerate} where $p(x)$ and $p(y)$ describe the distributions of two domains of images $X$ and $Y$. The first condition ensures that the generated images appear to come from the desired domains and is enforced by training two discriminators on $X$ and $Y$ respectively. The second condition ensures that the information about a source image is encoded in the generated image and is enforced by a \emph{cyclic consistency loss} of the form $||GFx - x || + ||FGy - y||$. The hope is that the information about the source image $x$ is encoded semantically into elements of the generated image $Fx$.

For our experiments, we trained a CycleGAN model on a ``maps'' dataset consisting of approximately 1,000 aerial photographs $X$ and 1,000 maps $Y$. The model trained for 500 epochs produced two maps $F :X \to Y $ and $G: Y \to X$ that generated realistic samples from these image domains.

\section{Hidden Information}
\begin{figure}
    \centering
    \begin{subfigure}{0.32\textwidth}
        \includegraphics[width=.98\linewidth,keepaspectratio=true]{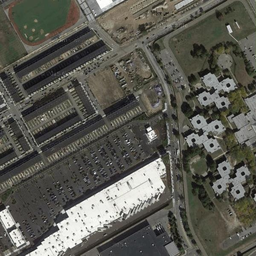}
        \caption{Aerial photograph: $x$.}
    \end{subfigure}
    \begin{subfigure}{0.32\textwidth}
        \includegraphics[width=.98\linewidth,keepaspectratio=true]{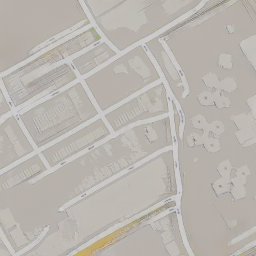}
        \caption{Generated map: $Fx$.}
    \end{subfigure}
    \begin{subfigure}{0.32\textwidth}
        \includegraphics[width=.98\linewidth,keepaspectratio=true]{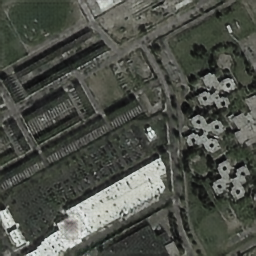}
        \caption{Aerial reconstruction: $GFx$.}
        \label{fig:aerial_rec}
    \end{subfigure}
    \caption{
        Details in $x$ are reconstructed in $GFx$, despite not appearing in the intermediate map $Fx$.
    }
    \label{fig:phenomenon}
\end{figure}

We begin with a curious observation, illustrated in Figure~\ref{fig:phenomenon}. We first take an aerial photograph $x$ that was unseen by the network at training time. Since the network was trained to minimize the cyclic consistency loss, one would expect that $x \approx GFx$, and indeed the two images turn out to be nearly identical. However, upon closer inspection, it becomes apparent that there are many details present in both the original aerial photograph $x$ and the aerial reconstruction $GFx$ that are not visible in the intermediate map $Fx$. For example, the pattern of black dots on the white roof in $x$ is perfectly reconstructed, even though in the map, that area appears solidly gray. How does the network know how to reconstruct the dots so precisely? We observe this phenomenon with nearly every aerial photograph passed into the network, as well as when CycleGAN is trained on datasets other than maps.

We claim that CycleGAN is learning an encoding scheme in which it ``hides'' information about the aerial photograph $x$ within the generated map $Fx$. This strategy is not as surprising as it seems at first glance, since it is impossible for a CycleGAN model to learn a perfect one-to-one correspondence between aerial photographs and maps, when a single map can correspond to a vast number of aerial photos, differing for example in rooftop color or tree location.

\begin{figure}
    \centering
    \begin{subfigure}{0.32\textwidth}
        \includegraphics[width=.98\linewidth,keepaspectratio=true]{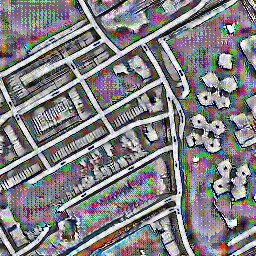}
        \caption{Generated map.}
    \end{subfigure}
    \begin{subfigure}{0.32\textwidth}
        \includegraphics[width=.98\linewidth,keepaspectratio=true]{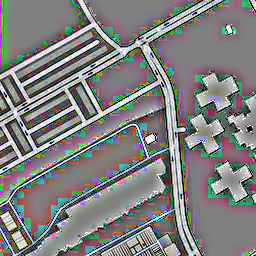}
        \caption{Training map, for comparison.}
    \end{subfigure}
    \caption{
        Maps with details amplified by adaptive histogram equalization. Information is present in the generated map even in regions that appear empty to the naked eye.
    }
    \label{fig:amplified}
\end{figure}

It may be possible to directly see where CycleGAN may be encoding this hidden information. When we zoom into an apparently solid region of the generated map, we in fact find a surprising amount of variation. We amplify this variation using an image processing technique called adaptive histogram equalization, which enhances contrast in a local neighborhood of each pixel, and present the results in Figure~\ref{fig:amplified}. For comparison, we apply the same transformation to a ground truth training map. We see that there does appear to be an extra, high-frequency signal in the generated map. We investigate the nature of this encoding scheme in the following sections.

\section{Sensitivity to Corruption}
\label{sec:sensitivity}
In this section, we corrupt the generated map with noise and study how the aerial reconstruction changes, in order to characterize the nature of the encoding scheme. Specifically, let us define
\begin{equation}
    V \equiv \E_{x \sim p(x),z\sim p(z)} || G (F x + z) - GF x ||_1,
\end{equation}
where $p(x)$ is the true aerial image distribution and $p(z)$ is a noise distribution. $V$ measures how different the aerial reconstruction is when noise is added, and we are interested in how $V$ depends on the noise distribution $p(z)$. In our experiments, we chose $p(z)$ to be Gaussian noise and varied the standard deviation $\epsilon$ and spatial correlation $\sigma$.

\begin{figure}
    \centering
    \begin{subfigure}{0.48\textwidth}
        \includegraphics[width=.98\linewidth,keepaspectratio=true]{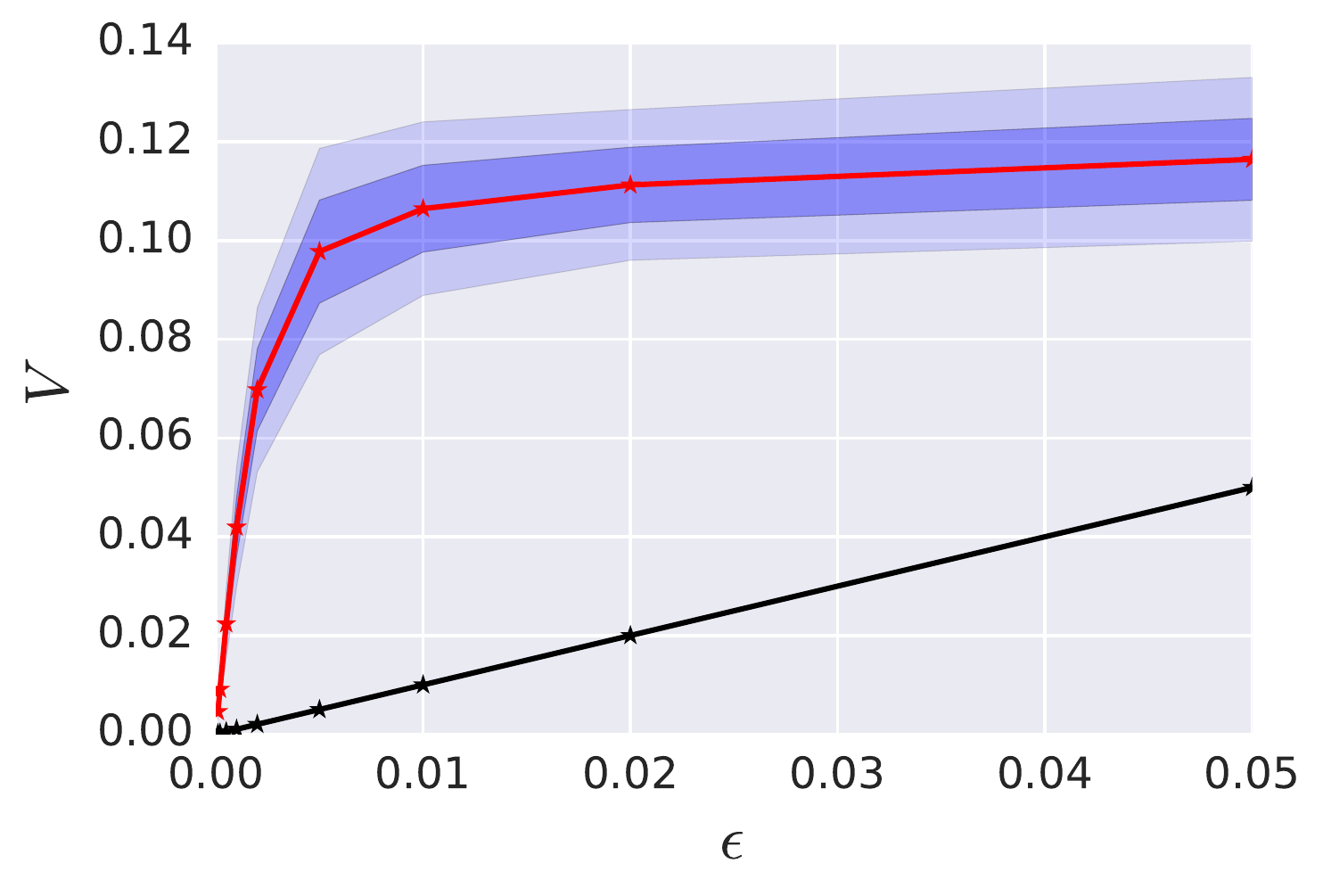}
        \caption{$V$ at $\sigma=0$ (spatially independent noise).}
    \end{subfigure}
    \begin{subfigure}{0.48\textwidth}
        \includegraphics[width=.98\linewidth,keepaspectratio=true]{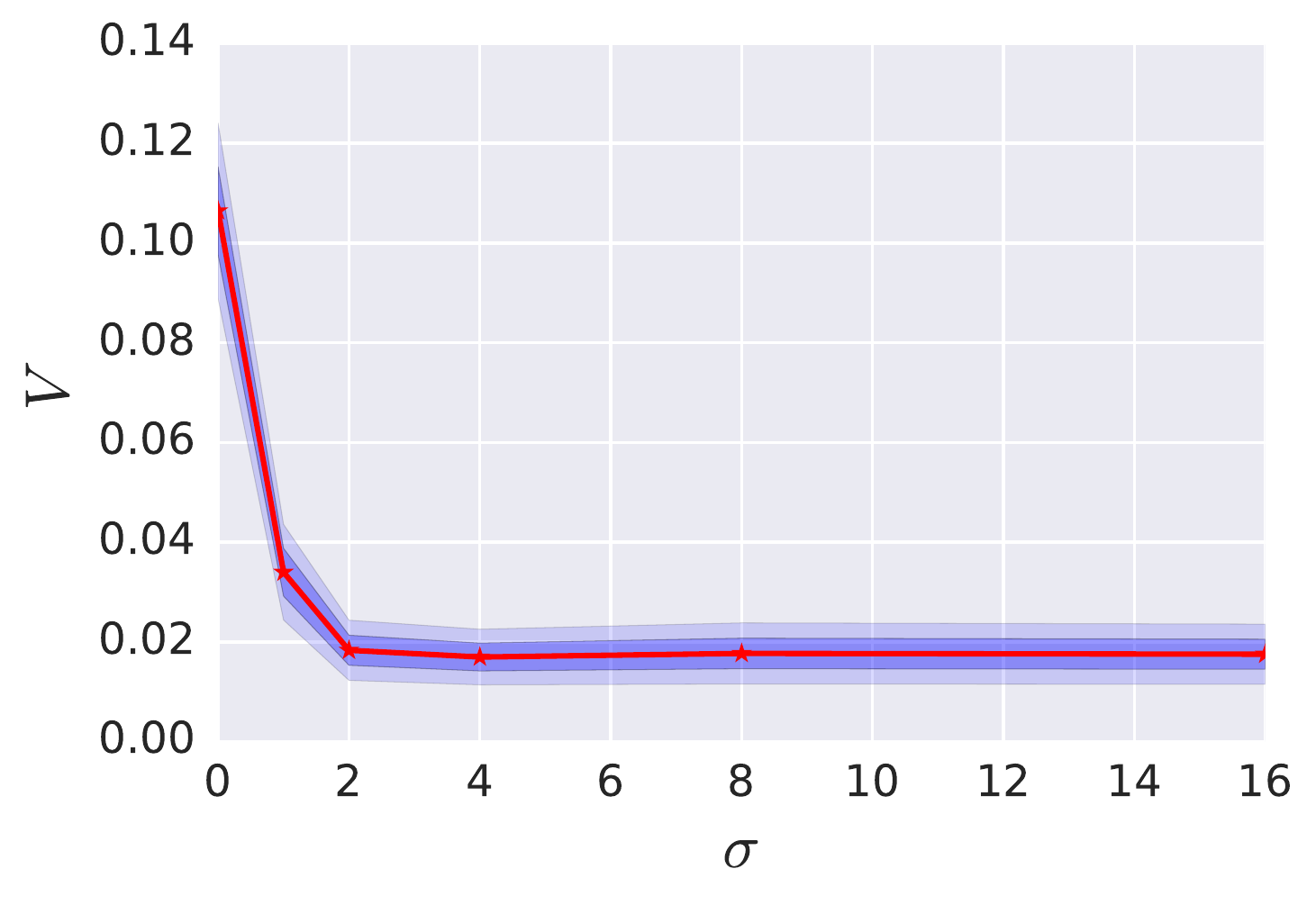}
        \caption{$V$ at $\epsilon=0.01$.}
    \end{subfigure}
    \caption{
        Sensitivity of $G$ to noise as the amplitude $\epsilon$ and spatial correlation $\sigma$ of the noise varies. In (a), $V = \epsilon$ is plotted for reference.
    }
    \label{fig:noise}
\end{figure}

Figure~\ref{fig:noise} depicts how $V$ behaves as a function of $\epsilon$ and $\sigma$, where the expectation is approximated by averaging over 50 aerial photographs. We found that $V$ attains nearly its maximum value as soon as $\epsilon \ge 3/256 \approx 0.01$, which corresponds to only 3 levels when the image is quantized by 8-bit integers. Thus an imperceptible modification of the map image can lead to major changes in the reconstructed aerial photograph. In fact, we found that simply encoding the generated map $Fx$ with lossy JPEG compression destroyed the reconstruction. We also found that $V$ quickly decays to its minimum value as soon as $\sigma \ge 2$, indicating that the information is fairly robust to low-frequency content -- including what we perceive as the map itself. This suggests that the majority of information about the source photograph is stored in a high-frequency, low-amplitude signal within the generated map.

\begin{figure}
    \centering
    \begin{subfigure}{0.23\textwidth}
        \includegraphics[width=.98\linewidth,keepaspectratio=true]{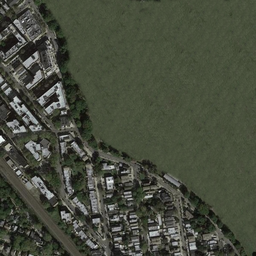}
        \caption{Original image: $x$.}
    \end{subfigure}
    \begin{subfigure}{0.23\textwidth}
        \includegraphics[width=.98\linewidth,keepaspectratio=true]{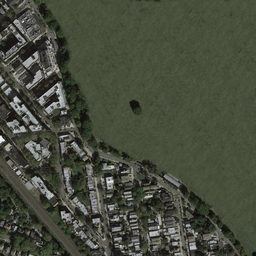}
        \caption{Edited image: $x'$.}
    \end{subfigure}
    \begin{subfigure}{0.23\textwidth}
        \includegraphics[width=.98\linewidth,keepaspectratio=true]{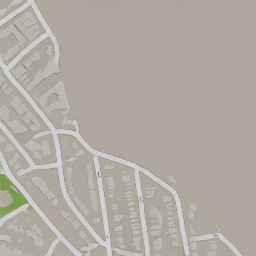}
        \caption{Generated map: $Fx'$.}
    \end{subfigure}
    \begin{subfigure}{0.23\textwidth}
        \includegraphics[width=.98\linewidth,keepaspectratio=true]{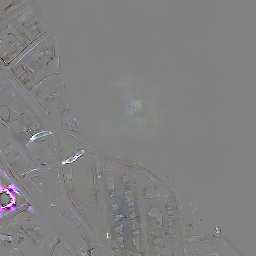}
        \caption{Difference: $Fx' - Fx$.}
        \label{fig:tree_delta}
    \end{subfigure}
    \caption{
        The information encoding is surprisingly non-local.
    }
    \label{fig:alter}
\end{figure}

While $G$ is quite sensitive to noise added to a map $Fx$, we show that $G$ is well-behaved when a perturbation $\Delta$ created by $F$ is added to $Fx$. Towards this end, we manually create two aerial images $x'$ and $x''$ by editing a tree onto a grass field in $x$ in two different locations; we then study the differences in the generated map $\Delta' = Fx'-Fx$ and $\Delta'' = Fx'' - Fx$, depicted in Figure~\ref{fig:alter}.\footnote{Interestingly, the change $\Delta'$ is not localized around the added tree but extends far beyond this region. However, the tree is still reconstructed when we mask out the nonlocal pixels.} We find that the reconstruction $G(Fx + \Delta' + \Delta'')$ contains both trees added in $x'$ and $x''$ and does not contain any unexpected artifacts. This may indicate that the encodings $\Delta'$ and $\Delta''$ are small enough that they operate in the linear regime of $G$, where \begin{equation}
    G(Fx + \Delta' + \Delta'') = GFx + dG \Delta' + dG \Delta'' + O(\Delta^2),
\end{equation} so that addition of perturbations in the map corresponds to independent addition of features in the generated aerial image. We confirmed numerically that $G(F x + \varepsilon\Delta')$ and $G(F x + \varepsilon\Delta'')$ are approximately linear with respect to $\varepsilon$. Finally, we note that if $\Delta'$ is added to an entirely different image, the generated aerial image does not necessarily reconstruct a tree and sometimes contains artifacts.

\section{Information Hiding as an Adversarial Attack}

\begin{figure}
    \centering
    
    \begin{subfigure}{0.23\textwidth}
        \includegraphics[width=.98\linewidth,keepaspectratio=true]{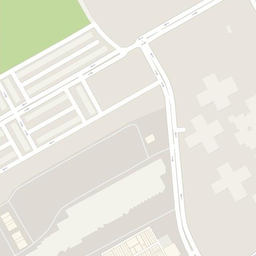}
    \end{subfigure}
    \begin{subfigure}{0.23\textwidth}
        \includegraphics[width=.98\linewidth,keepaspectratio=true]{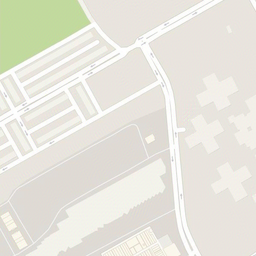}
    \end{subfigure}    
    \begin{subfigure}{0.23\textwidth}
        \includegraphics[width=.98\linewidth,keepaspectratio=true]{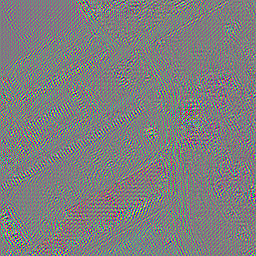}
    \end{subfigure}
    \begin{subfigure}{0.23\textwidth}
        \includegraphics[width=.98\linewidth,keepaspectratio=true]{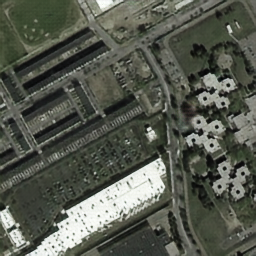}
    \end{subfigure}
    
    \begin{subfigure}{0.23\textwidth}
        \includegraphics[width=.98\linewidth,keepaspectratio=true]{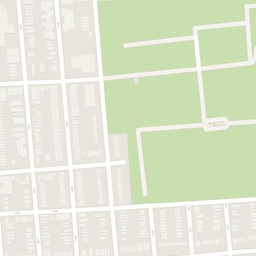}
        \caption{Source map: $y_0$.}
    \end{subfigure}
    \begin{subfigure}{0.23\textwidth}
        \includegraphics[width=.98\linewidth,keepaspectratio=true]{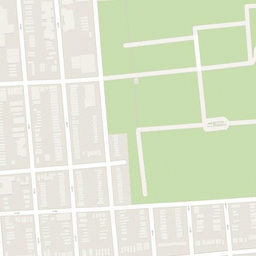}
        \caption{Crafted map: $y^*$.}
    \end{subfigure}    
    \begin{subfigure}{0.23\textwidth}
        \includegraphics[width=.98\linewidth,keepaspectratio=true]{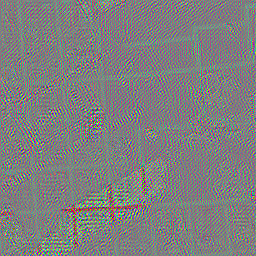}
        \caption{Difference: $y^* - y_0$.}
        \label{fig:map_delta}
    \end{subfigure}
    \begin{subfigure}{0.23\textwidth}
        \includegraphics[width=.98\linewidth,keepaspectratio=true]{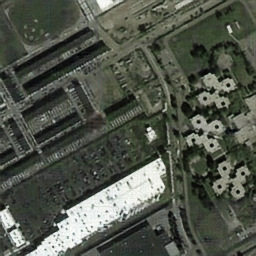}
        \caption{Reconstruction: $Gy^*$.}
    \end{subfigure}    
    \caption{
        Generation of a single target aerial photo $x^*$ from two arbitrary maps $y_0$. Note that (c) is amplified for visibility.
    }
    
    \label{fig:maps}
\end{figure}

In this section, we demonstrate that $G$ has the ability to reconstruct \emph{any} desired aerial photograph $x^*$ from a specially crafted map $y^*$. Specifically, we solve the optimization problem \begin{equation}
    y^* = \argmin_{y} || Gy - x^* ||, \label{eq:adversarial}
\end{equation} by starting gradient descent from an initial, source map $y_0$. This is in the same spirit as an adversarial attack \citep{szegedy2013intriguing} on $G$, where we are constructing an input $y$ that forces $G$ to produce a desired photograph $x^*$. We present the results in Figure~\ref{fig:maps}; we find that the addition of a low-amplitude signal to virtually any initial map $y_0$ is sufficient to produce a given aerial image, and the specially crafted map $y^*$ is visually indistinguishable from the original map $y_0$. The fact that the required perturbation is so small is not too surprising in light of Section~\ref{sec:sensitivity}, where we showed that tiny perturbations in $G$'s input result in large changes in its output.

Recognizing that the cyclic consistency loss \[\argmin_{F,G} || GFx - x || \] is similar in form to the adversarial attack objective in Equation~\ref{eq:adversarial}, we may view the CycleGAN training procedure as continually mounting an adversarial attack on $G$, by optimizing a generator $F$ to generate adversarial maps that force $G$ to produce a desired image. Since we have demonstrated that it is possible to generate these adversarial maps using gradient descent, it is nearly certain that the training procedure is also causing $F$ to generate these adversarial maps. As $G$ is also being optimized, however, $G$ may actually be seen as cooperating in this attack by learning to become increasingly susceptible to attacks. We observe that the magnitude of the difference $y^* - y_0$ necessary to generate a convincing adversarial example by Equation~\ref{eq:adversarial} decreases as the CycleGAN model trains, indicating cooperation of $G$ to support adversarial maps.

\section{Discussion}

CycleGAN is designed to find a correspondence between two probability distributions on domains $X$ and $Y$. However, if $X$ and $Y$ are of different complexity -- if their distributions have differing entropy -- it may be impossible to learn a one-to-one transformation between them. We demonstrated that CycleGAN sidesteps this asymmetry by hiding information about the input photograph in a low-amplitude, high-frequency signal added to the output image.

By encoding information in this way, CycleGAN becomes especially vulnerable to adversarial attacks; an attacker can cause one of the learned transformations to produce an image of their choosing by perturbing any chosen source image. The ease with which adversarial examples may be generated for CycleGAN is in contrast to previous work by \cite{tabacof2016adversarial} and \cite{kos2017adversarial}, which illustrate that the same attack on VAEs requires noticeable changes to the input image. In serious applications, the cyclic consistency loss should be modified to prevent such attacks. In a future work, we will explore one possible defense: since this particular vulnerability is caused by the cyclic consistency loss and the difference in entropy between the two domains, we investigate the possibility of artificially increasing the entropy of one of the domains by adding an additional hidden variable. For instance, if a fourth image channel is added to the map, information need not be hidden in the image but may instead be stored in this fourth channel, thus reducing the need for the learned transformations to amplify their inputs and making the possibility of attack less likely.

The phenomenon also suggests one possible route for improving the quality of images generated by CycleGAN. Even though the cyclic consistency loss is intended to force the network into encoding information about a source image semantically into the generated image, the model in practice learns to ``cheat'' by encoding information imperceptibly, adversarially. If the network were somehow prevented from hiding information, the transformations may be forced to learn correspondences that are more semantically meaningful.

More broadly, the presence of this phenomenon indicates that caution is necessary when designing loss functions that involve compositions of neural networks: such models may behave in unintuitive ways if one component takes advantage of the ability of the other component to support adversarial examples. Common frameworks such as generative adversarial networks (\cite{goodfellow2014generative}) and perceptual losses (e.g.~\cite{johnson2016perceptual}) employ these compositions; these frameworks should be carefully analyzed to make sure that adversarial examples are not an issue.

\subsubsection*{Acknowledgments}
We thank Jascha Sohl-Dickstein for his insightful comments.

\nocite{*}
\bibliographystyle{abbrvnat}
\bibliography{references}

\end{document}